# Optimal Hyperparameters and Structure Setting of Multi-Objective Robust CNN Systems via Generalized Taguchi Method and Objective Vector Norm

Sheng-Guo Wang* and Shanshan Jiang

University of North Carolina at Charlotte, Charlotte, USA
swang@uncc.edu, ssjiang1991@gmail.com;   *Corresponding Author

**Abstract:** Recently, Machine Learning (ML), Artificial Intelligence (AI), and Convolutional Neural Network (CNN) have made huge progress with broad applications, where their systems have deep learning structures and a large number of hyperparameters that determine the quality and performance of the CNNs and AI systems. These systems may have multi-objective ML and AI performance needs. There is a key requirement to find the optimal hyperparameters and structures for multi-objective robust optimal CNN systems. This paper proposes a generalized Taguchi approach to effectively determine the optimal hyperparameters and structure for the multi-objective robust optimal CNN systems via their objective performance vector norm. The proposed approach and methods are applied to a CNN classification system with the original ResNet for CIFAR-10 dataset as a demonstration and validation, which shows the proposed methods are highly effective to achieve an optimal accuracy rate of the original ResNet on CIFAR-10.

## 1  Introduction

Artificial Intelligence (AI) research has been spread into broadly areas with various applications, e.g., for robots, traffic, medical image, protein structure [1] - [3] where Convolutional Neural Network (CNN) plays an important role for image recognition, medical diagnosis, environmental identification, tracking and analysis. The AI systems and CNN models have their various deep structures with a large number of hyperparameters. These hyperparameters and structures significantly determine the performance and quality of the CNNs and AI systems.

Thus, there is an important optimization issue in AI systems and CNNs. To address that, some research took the ways for progress by architecture search and random tuning hyperparameters, e.g., auto hyperparameter and architecture search (AutoHAS) by linear basis combination to extend the weight sharing in 2020 [4], random search based on grid search in 2011 [5], and the Bayesian method for a global optimum [6]. However, there are still problems as the convergence issue to global optimality in random search and the sensitive problem in the Bayesian search.

On the other hand, Wang [7] first proposed to apply the Taguchi's orthogonal test design method [8] for the machine learning (ML) to choose optimal hyperparameters of multi-variable resolutions in tidal wetland modeling and prediction in 2014. Then, Wang *et. al*. reported their research method and results on the best resolution parameters setting for key variables in machine learning Random Forest (RF) model and Logit model via the Taguchi approach in 2018 [9]. Zhang et al. [10] [11] first used Taguchi orthogonal array tuning method (OATM) to tune their deep learning hyperparameters of RNN (Recurrent Neural Networks) and CNN structures for their applications, e.g., EEG signals, in 2019. They compared the OATM to grid search optimization, random search optimization and Bayesian optimization methods under the same nine steps search, where the OATM can significantly save the tuning time while preserving the satisfying performance. Recently, Munger and Desa [12] applied the Taguchi method to search better models among RF, SVM (supporting vector machine) and Naïve Bayes models in 2021.

However, there is still a lack of research on hyperparameters and structures setting for multi-objective AI systems and CNN systems. This is the research motivation of this paper. We propose new methods to search optimal hyperparameters and structure for multi-objective robust optimal AI CNN systems via our generalized Taguchi method and the multi-objective performance functional vector norm. The new methods are applied to searching an optimal CNN classification system for the image classification benchmark dataset CIFAR-10 [13] to validate our proposed approach and methods. The experimental results show that our proposed new methods can fast set optimal hyperparameters and structure for single-objective and multi-objective optimal AI CNN systems to achieve a best accuracy rate of a well-known CNN backbone ResNet [14] on the CIFAR-10 test dataset using its original ResNet type layers in the literature by 16 systematic orthogonal tests.

## 2  Multi-Objective AI CNN System and its Optimal Performance Functional

The goal of this paper is to build a robust optimal AI CNN system model with its multi-objective, where the single objective is a special case of the multi-objective. In this section, we present the system optimal multi-objective and its performance functional vector norm with its performance functional value scaling and normalization.

Let a multi-objective performance functional of a CNN system be a vector functional with its $m$ components, each one for one objective respectively, as

$$P = [p_j, j = 1, \cdots, m] \tag{1}$$

where its functional parameter set $H$ include CNN system hyperparameters and structures as

$$P = P(H) = [p_j(h_k, k = 1, \cdots, K), j = 1, \cdots, m], \quad H = [h_k, k \in \{1, \cdots, K\}]. \tag{2}$$

Since the values of each objective functional $p_j(H)$ may be at different levels and in a different range, thus we will do a normalization process for each $p_j(H)$ by two steps. The first step is a scaling process to scale each $p_j(H)$ such that $p_j(H), j = 1, \cdots, m$, are in a similar suitable level and range, by their respective scale functional $s_j(p_j(H))$. The second step is a weighting process to weight each scaled performance functional $s_j(p_j(H))$ by a weighting factor $\alpha_j$ based on its objective importance in the multi-objective family, where the sum of $m$ weighting factors is one. We denote the normalized performance functional vector as $\bar{P}$

$$\bar{P} = [\bar{p}_j, j = 1, \cdots, m] = [\alpha_j s_j(p_j(H)), j = 1, \cdots, m], \quad \sum_{j=1,\cdots,m} \alpha_j = 1 \tag{3}$$

Then, we propose to take the norm of the multi-objective performance functional vector as our multi-objective performance functional $J$ for optimization as

$$J(\bar{P}) = norm(\bar{P}) \tag{4}$$

where the norm may be any norm of the normalized functional vector for consideration. Now the goal is to find the optimal hyperparameters $H^*$(which may include the structure parameters and/or the structure types) to minimize (or maximize) our multi-objective performance functional $J$ for the optimal $J^*$ as

$$J^* = \min_H J(\bar{P}(H)) = J(\bar{P}(H^*)), \quad H^* = \arg\min_H J(\bar{P}(H)) = arg \min_{h_k, k \in \{1,\cdots,K\}} J(\bar{P}(H)). \tag{5}$$

We may take any norm for our multi-objective performance vector as a performance measurement. In this paper, we take 2-norm $J_{2-norm}$ as the performance index $J$, i.e., the multi-objective performance vector length and distance, as

$$J_{2-norm} = \left(\sum_{j=1,\cdots,m} \bar{p}_j^2\right)^{\frac{1}{2}} = \left(\sum_{j=1,\cdots,m} (\alpha_j s_j(p_j))^2\right)^{\frac{1}{2}} \tag{6}$$

where $\bar{p}_j, j = 1, \cdots, m$, are set as positive functionals of the parameter $H = [h_k, k = 1, \cdots, K]$.

For single objective case, it is a special case of the multi-objective case with

$$m = 1, \quad \alpha_j = \alpha_1 = \alpha = 1, \quad s_j = s_1 = s, \quad \bar{P} = \bar{p}, \quad P = p, \tag{7}$$

where the functional $s(\cdot)$ may be a simple identity function, i.e., $s(p) = p$. For classification and identification system, its single objective is the system accuracy, i.e., to minimize an error performance functional $J$. For multi-objective systems, the system accuracy will still be one of the most important objectives.

When we consider a multi-objective optimization problem, there are some methods in the literature, e.g., to satisfy a set of individual objective performance criterial indices of a multi-objective system [15] [16]. However, here we present our new approach and method via the normalization, objective performance vector norm and generalized Taguchi method for AI CNN systems as described further below.

## 3  Hyperparameters and Structure

In this section, we present the generalized hyperparameters, which include the normal hyperparameters, the system structure types and their parameters. These general hyperparameters determine the CNN system performance. In our experiments, we select the following five general hyperparameters:

$h_1 = initial\ learning\ rate\ (lr),\ h_2 = epoch\ number\ (epoch),\ h_3 = training\ sampling\ rate\ (sampling)$
$h_4 = backbone\ ResNet\ (backbone)\ layers,\ \ \ h_5 = batch\ size\ (batch).$

## 4 Methods via Extended Taguchi Approach

Taguchi method has been successfully applied to various technical and industrial areas by using orthogonal experiments to search and set the optimal system parameters for decades. However, it is just the recent years to be applied to ML and AI areas. The common Taguchi method utilizes the signal-to-noise ratio as its objective performance function for optimization. Here, we extend the method to multi-objective performance functional vector norm for optimization, and let the system variables be the generalized hyperparameters including the system structure types and/or system parameters in addition to normal hyperparameters. Thus, we call our proposed method the extended general Taguchi method. We further briefly describe the extended Taguchi method below.

**Step 1:** Define the optimal objective index functional $J$ for a single-objective or multi-objective system, e.g., as its performance vector norm as in (4).

**Step 2:** Determine the general hyperparameter set $H = [h_1, \cdots, h_k, \cdots, h_K]$, where the objective performance functional $J(H)$ is a functional of $H$.

**Step 3:** Design the grid level number $L_k$ for each parameter $h_k$, and its grid values $h_{k,l_k}$, where $k$ is for the k-th parameter, $1 \le k \le K$, and $h_{k,l_k}$ is the $l_k$-th grid value of $h_k$, $1 \le l_k \le L_k$. Thus, the number of total grid points is

$$N = \prod_{k=1,\cdots,K} L_k. \tag{8}$$

**Step 4:** Design an orthogonal experimental table $T$ for $K$ parameters with a max $L_M$ level. Table $T$ can be found in [17] after $K$ and $L_M$ are determined. For simplicity, we assume each $L_k$ is the same as $L$ as $L_M$.

**Step 5:** Run $R$ orthogonal experiments with the designed orthogonal parameter level value set in Step 3 as the Table $T$ listed levels, where $R$ is the orthogonal test number. For example, we have $K = 5$ and $L = 4$ in our experiments, thus we have table $T = T_{16}(L_4^5)$ for the above selected five parameters in Section 4 as Table 1 below, where $R = 16$ is the orthogonal test number.

**Table 1. Taguchi 16 orthogonal experiment table** (L16_4 [17])

| Experiment # | Hyperparameter Levels | | | | |
|---|---|---|---|---|---|
| | learning rate - lr | epochs | sampling | Backbone ResNet | Batch-size |
| 0 | 1 | 4 | 4 | 4 | 4 |
| 1 | 2 | 3 | 4 | 1 | 2 |
| 2 | 4 | 1 | 4 | 2 | 3 |
| 3 | 1 | 1 | 1 | 1 | 1 |
| 4 | 2 | 4 | 3 | 2 | 1 |
| 5 | 2 | 1 | 2 | 3 | 4 |
| 6 | 4 | 3 | 2 | 4 | 1 |
| 7 | 4 | 2 | 3 | 1 | 4 |
| 8 | 3 | 2 | 4 | 3 | 1 |
| 9 | 3 | 1 | 3 | 4 | 2 |
| 10 | 1 | 3 | 3 | 3 | 3 |
| 11 | 4 | 4 | 1 | 3 | 2 |
| 12 | 3 | 3 | 1 | 2 | 4 |
| 13 | 1 | 2 | 2 | 2 | 2 |
| 14 | 2 | 2 | 1 | 4 | 3 |
| 15 | 3 | 4 | 2 | 1 | 3 |

**Step 6:** Analyze the orthogonal experiment performance functional $J$. For each level setting $h_{k,l_k}$ of parameter $h_k$ in the orthogonal table $T$, we take its average performance functional. For example, we take the average performance functional value of its experiments (0, 3, 10, 13) as a group $(k, l_k) = (1,1)$ for parameter $h_1$ at its level 1 value, then experiments (1, 4, 5, 14) as a group $(k, l_k) = (1,2)$, experiments (8, 9, 12, 15) as a group $(k, l_k) = (1,3)$, and experiments (2, 6, 7, 11) as a group $(k, l_k) = (1,4)$. Then, we select the corresponding parameter level value with the minimum average group performance functional value among these four groups as the suggested optimal parameter value of this parameter $h_1$. For example, if group (1,4) has the minimum average group performance value among four groups (1,1), (1,2), (1,3) and (1,4), then we take the level 4 value $h_{1,4}$ of $h_1$ as the optimal value $h_1^*$ of parameter $h_1$. Similarly, we find the optimal parameter grid value $h_k^*$ for each $h_k$, $k = 1, \cdots, 5$, respectively.

**Step 7:** Get the suggested optimal parameter setting $H^* = [h_1^*, \cdots, h_K^*]$ in the design parameter grid set from the above analysis Step 6 as in (5).

**Step 8:** Run the suggested optimal model test with the suggested optimal parameter setting $H^*$ and get its performance functional $J(H^*)$ for verification, i.e.,

$$J(H^*) = J(h_k^*, k = 1, \cdots, K). \tag{9}$$

The extended Taguchi orthogonal experiment table has total experiments $R = 1 + K(L-1)$ [18] that is much less than the total possible grid points number $N$ in (8). In the above example, $R = 1 + 5 * (4-1) = 16 < N = 4^5 = 1024$, i.e., a more than 1000 experiments saved.

From Step 6 for each parameter $h_k$, we can calculate its average group performance functional variation range among these level setting $h_{k,l_k}, 1 \leq l_k \leq L_k$ in the orthogonal table $T$. Then, the variation range value of parameter $h_k$ reflects the importance rank of parameter $h_k$ to the performance variation among these $K$ parameters.

## 5 Orthogonal Experiments

This section describes the experiments to apply our methods to a CNN classification system on CIFAR-10 dataset. The experiments include two objective cases: (1) a single objective case for the system accuracy, and (2) as a multi-objective example, a bi-objective case for the system accuracy and training speed by using the performance functional vector 2-norm. In the learning process, we use the stochastic gradient descent (SGD) optimizer (batch gradient descent) with a momentum of 0.9 and a weight decay of 1e-4. For the training speed, we consider a flat random sampling rate as in [19], which is the factor $h_3$ here. For the CNN system backbone, we take the popular ResNet [14], which is broadly used in different computer vision tasks. We use the Pytorch framework ResNet implementation, which refers to [20], on two Nvidia Quadro RTX 5000 GPUs.

We select five general parameters and each parameter has four levels, i.e., $K = 5, k = 1, \cdots, 5, L_k = L = 4, l_k = 1, \cdots, 4$. Their level values are as follows.

(i)   Initial Learning rate $h_1$:        0.01,   0.03,   0.05,   0.1;
(ii)  Epoch number $h_2$:                 60,     90,     120,    150;
(iii) Training sampling size $h_3$:       0.382,  0.618,  0.8,    1.0;
(iv)  Backbone ResNet layers $h_4$:       20,     32,     56,     110;
(v)   Batch size $h_5$:                   32,     64,     128,    256.

Among above five hyperparameters except the first one, all other four are first chosen for the Taguchi method applied to CNN system in the literature. The learning rate starts at an initial rate level of $h_1$, and then it is multiplied by 0.1 and 0.01 at 50% and 75% of the epochs, respectively. The above four training sampling size levels of $h_3$ are selected as full sampling, 80-20 rule, optimization golden ratio 0.618, and another end 0.382, where the fast sampling rate of 80-20 rule and optimization golden ration 0.618 refers to [19].

For case (1), the single objective is as objective 1 to minimize the system error rate $e$. Its normalization functional is just the identity functional, so we have

$$\bar{P} = P = \bar{p} = p = e, \ \alpha s(e) = e, \ J(\bar{P}) = J(P) = e. \tag{10}$$

In case (2), the multi-objective is a bi-objective taking the system error rate $e$ and the system training time $t$. We take the second objective as the training time because the training process of AI systems takes a huge time in the deep learning, which becomes an issue in applications, e.g., in limited resource cases, mobile systems and fast systems.

The bi-objective case has the 2-norm of its normalized performance functional vector $\bar{P}_2$ as $J(\bar{P}_2)$

$$J(\bar{P}_2) = \sqrt{(\bar{e})^2 + (\bar{t})^2} = \sqrt{(\alpha_e \cdot e)^2 + (\alpha_t \cdot s_2(t))^2} \tag{11}$$

$$\bar{e} = \alpha_e \cdot e, \ s_2(t) = \log(t)/1000, \ \bar{t} = \alpha_t \log(t)/1000, \ \alpha_e = 0.8, \ \alpha_t = 1 - \alpha_e = 0.2. \tag{12}$$

Based on the selection of five general hyperparameters and four levels, i.e., $K = 5$ and $L = 4$, we have the Taguchi orthogonal table with 16 orthogonal experiments as listed in above Table 1 instead of 1024 experiments.

### 5.1. Training Dataset and Test Dataset

We select the well-known CIFAR-10 dataset [13] for our experiments and test. The CIFAR-10 dataset contains 10 classes for identification, and has 50k images in training set for model training and 10k images in test set for test evaluation.

The common Taguchi method takes one dataset to search its optimal parameter set for the system or process by analyzing the orthogonal experiments and evaluating the searched system performance under this same dataset, i.e., there is only one dataset for training, analyzing, searching and evaluating the optimal system. However, in ML, DL, AI and CNN we usually have two different datasets, a training dataset and a test dataset. Thus, we extend the common Taguchi method to the extended Taguchi method for two datasets, i.e., one training dataset and another test dataset. The former is for training, analyzing and searching the optimal hyperparameter set of systems, and the latter is for evaluating the systems performance, where they are different datasets for more practical meanings, although there are some papers which may still use one dataset for both training and testing in ML, DL and AI, e.g., those in [10] [11].

**Comment 1 (on Steps 5 and 6, Section 4):** The system and its functional $J$ are trained and analyzed on the training set for searching an optimal system, while the searched optimal system and its experimental systems are finally evaluated on the test set, which is not used for the system training as two separated data sets in CIFAR-10. The traditional Taguchi method uses only one data set for training, analysis, search and evaluation. Thus, it is a further extension as our extended Taguchi approach to check the built system optimality and robustness.

### 5.2. Experimental Results

As listed in Table 1, we run16 experiments on the CIFAR-10 dataset for two objective cases as above described. The systems are trained and analyzed on the training set and tested on the test set, but are evaluated on both training set and test set for robustness check. Finally, the searched optimal system is compared with the 16 orthogonal experimental systems in Table 2. All experiments are run on a Lenovo workstation with two Nvidia Quadro RTX 5000 GPUs.

For concise reading, we list these two objective cases on the training and test sets together with their experimental data in Table 2, where the suggested optimal hyperparameters and system model performance by the proposed methods are also listed in the last two rows of Table 2 for comparison.

Let us describe experiment 0 as an example for all other experiments as follows. In experiment index 0, training error is $e = 0.0023 = 0.23\%$ and its training time is $t = 3284.64$ sec, where the training time is for the model training time only, not including evaluation time and test time after the training. The "training obj. 1" value is exactly the "training error $e$" 0.0023 for the single objective case.

For the bi-objective case, we first take a scaling process for the training time $t$ in the normalization as $s_2(t) = \log(t)/1000$ such that their values are in a similar range comparable to another objective-error-rate range for a meaningful weighting process. For experiment 0, $s_2(t) = \log(3284.64)/1000 = 0.0035$, that is in a similar range of $s_1(e) = e = 0.0023$, as we observe the columns "training error e" and "log(t)/1000".

Thus, we have "training obj. 1" and "training obj.2" values calculated by (10) and (11) respectively as follows:

$$\text{Training Obj. 1} = J(P) = P = P_1 = p_1 = e = 0.0023,$$

$$\text{Training Obj.2} = J(\bar{P}_2) = \sqrt{(\alpha_e \cdot e)^2 + (\alpha_t \cdot s_2(t))^2} = \sqrt{(0.8 * 0.0023)^2 + (0.2 * \log(3284.64)/1000)^2} = 0.0020,$$

where $\alpha_e = 0.8, \alpha_t = 0.2$. For "test obj.1" and "test obj.2", we calculate their values by using (10) and (11), respectively, but based on their test accuracy error rate on the test set as listed in column "Test Obj.1" and the training time as listed in the same column "Training time t (sec)".

**Table 2. Experiment results on CIFAR-10 trained by the training set and tested by the test set**

| Exp. index | Learn. Rate | Epochs | Sample Rate | Str. ResNet | Batch Size | Training Error e | Training time t (sec) | Log(t)/1000 | Training Obj. 1 | Training Obj.2 | Test Obj. 1 | Test Obj. 2 |
|---|---|---|---|---|---|---|---|---|---|---|---|---|
| 0 | 0.01 | 150 | 1 | 110 | 256 | 0.0023 | 3284.64 | 0.0035 | 0.0023 | 0.0020 | 0.1144 | 0.0915 |
| 1 | 0.03 | 120 | 1 | 20 | 64 | 0.0087 | 3253.44 | 0.0035 | 0.0087 | 0.0070 | 0.0863 | 0.0690 |
| 2 | 0.1 | 60 | 1 | 32 | 128 | 0.0144 | 993.68 | 0.0030 | 0.0144 | 0.0115 | 0.0863 | 0.0690 |
| 3 | 0.01 | 60 | 0.382 | 20 | 32 | 0.0967 | 1262.41 | 0.0031 | 0.0967 | 0.0773 | 0.1357 | 0.1086 |
| 4 | 0.03 | 150 | 0.8 | 32 | 32 | 0.0024 | 8034.14 | 0.0039 | 0.0024 | 0.0021 | 0.0726 | 0.0581 |
| 5 | 0.03 | 60 | 0.618 | 56 | 256 | 0.0679 | 423.81 | 0.0026 | 0.0679 | 0.0544 | 0.1297 | 0.1038 |
| 6 | 0.1 | 120 | 0.618 | 110 | 32 | 0.0020 | 10272.0 | 0.0040 | 0.0020 | 0.0018 | 0.0659 | 0.0527 |
| 7 | 0.1 | 90 | 0.8 | 20 | 256 | 0.0240 | 535.30 | 0.0027 | 0.0240 | 0.0192 | 0.0971 | 0.0777 |
| 8 | 0.05 | 90 | 1 | 56 | 32 | 0.0017 | 7832.48 | 0.0039 | 0.0017 | 0.0015 | 0.0688 | 0.0550 |
| 9 | 0.05 | 60 | 0.8 | 110 | 64 | 0.0113 | 3365.06 | 0.0035 | 0.0113 | 0.0091 | 0.0808 | 0.0646 |
| 10 | 0.01 | 120 | 0.8 | 56 | 128 | 0.0099 | 2121.64 | 0.0033 | 0.0099 | 0.0079 | 0.1045 | 0.0836 |
| 11 | 0.1 | 150 | 0.382 | 56 | 64 | 0.0102 | 2469.93 | 0.0034 | 0.0102 | 0.0082 | 0.0787 | 0.0630 |
| 12 | 0.05 | 120 | 0.382 | 32 | 256 | 0.0452 | 414.25 | 0.0026 | 0.0452 | 0.0361 | 0.1126 | 0.0901 |
| 13 | 0.01 | 90 | 0.618 | 32 | 64 | 0.0373 | 1812.48 | 0.0033 | 0.0373 | 0.0299 | 0.1080 | 0.0864 |
| 14 | 0.03 | 90 | 0.382 | 110 | 128 | 0.0434 | 1227.94 | 0.0031 | 0.0434 | 0.0347 | 0.1097 | 0.0878 |
| 15 | 0.05 | 150 | 0.618 | 20 | 128 | 0.0147 | 1359.75 | 0.0031 | 0.0147 | 0.0118 | 0.0960 | 0.0768 |
| obj1 | 0.1 | 150 | 1 | 110 | 64 | 0.0001 | 10389.3 | 0.0040 | **0.0001** | 0.0008 | **0.0583** | 0.0466 |
| obj2 | 0.1 | 150 | 1 | 110 | 64 | 0.0001 | 10389.3 | 0.0040 | 0.0001 | **0.0008** | 0.0583 | **0.0466** |

Now, we analyze and identify the suggested optimal general hyperparameter set for the optimal system model from Table 2 based on Steps 6 and 7 in Section 4. For concise reading, we list the average group analysis for parameter $h_1$ in Table 3 as an example, where $h_1 = 0.1$ is the optimal $h_1^*$ based on the minimum average group performance according to the level $h_{1,4}$ of parameter Learning Rate $h_1$.

**Table 3. Average group performance analysis according to the levels of parameter 1 for Learning Rate $h_1^*$**

| Exp. Index | Learn Rate $h_1$ | Training error | Train. time t | Log(t)/1000 | Training Obj 1 | Training Obj 2 | average training obj1 | average training obj 2 |
|---|---|---|---|---|---|---|---|---|
| 0 | 0.01 | 0.0023 | 3284.6 | 0.0035 | 0.0023 | 0.0020 | 0.0366 | 0.0293 |
| 3 | 0.01 | 0.0967 | 1262.4 | 0.0031 | 0.0967 | 0.0773 | | |
| 10 | 0.01 | 0.0099 | 2121.6 | 0.0033 | 0.0099 | 0.0079 | | |
| 13 | 0.01 | 0.0373 | 1812.5 | 0.0033 | 0.0373 | 0.0299 | | |
| 1 | 0.03 | 0.0087 | 3253.4 | 0.0035 | 0.0087 | 0.007 | 0.0366 | 0.0293 |
| 4 | 0.03 | 0.0024 | 8034.1 | 0.0039 | 0.0024 | 0.0021 | | |
| 5 | 0.03 | 0.0679 | 423.8 | 0.0026 | 0.0679 | 0.0544 | | |
| 14 | 0.03 | 0.0434 | 1227.9 | 0.0031 | 0.0434 | 0.0347 | | |
| 8 | 0.05 | 0.0017 | 7832.5 | 0.0039 | 0.0017 | 0.0015 | 0.0182 | 0.0146 |
| 9 | 0.05 | 0.0113 | 3365.1 | 0.0035 | 0.0113 | 0.0091 | | |
| 12 | 0.05 | 0.0452 | 414.2 | 0.0026 | 0.0452 | 0.0361 | | |
| 15 | 0.05 | 0.0147 | 1359.8 | 0.0031 | 0.0147 | 0.0118 | | |
| 2 | 0.1 | 0.0144 | 993.7 | 0.0030 | 0.0144 | 0.0115 | 0.0127 | 0.0102 |
| 6 | 0.1 | 0.0020 | 10272 | 0.0040 | 0.0020 | 0.0018 | | |
| 7 | 0.1 | 0.0240 | 535.3 | 0.0027 | 0.0240 | 0.0192 | | |
| 11 | 0.1 | 0.0102 | 2469.9 | 0.0034 | 0.0102 | 0.0082 | | |

Similarly, we derive Tables 4 – 7 for hyperparameters $h_2, \cdots, h_5$, respectively, from Table 2 for their parameter level analysis and their optimal $h_2^*, \cdots, h_5^*$. We have the suggested optimal hyperparameter settings as follows,

$$H^* = \{h_1^*, h_2^*, h_3^*, h_4^*, h_5^*\} = \{0.1, 150, 1, 110, 64\} \qquad (13)$$

From Tables 3 – 7, this $H^*$ set in (13) is also for both two objective cases as listed at the last two rows in Table 2.

Table 4. Average group performance analysis according to the levels of parameter 2 for Epochs $h_2^*$

| Exp. Index | Epochs $h_2$ | Training error | Train. time t | Log(t)/1000 | Training Obj 1 | Training Obj 2 | average training obj1 | average training obj 2 |
|---|---|---|---|---|---|---|---|---|
| 3 | 60 | 0.0967 | 1262.4 | 0.0031 | 0.0967 | 0.0773 | 0.0476 | 0.0381 |
| 5 | 60 | 0.0679 | 423.8 | 0.0026 | 0.0679 | 0.0544 | | |
| 9 | 60 | 0.0113 | 3365.1 | 0.0035 | 0.0113 | 0.0091 | | |
| 2 | 60 | 0.0144 | 993.7 | 0.0030 | 0.0144 | 0.0115 | | |
| 13 | 90 | 0.0373 | 1812.5 | 0.0033 | 0.0373 | 0.0299 | 0.0266 | 0.0213 |
| 14 | 90 | 0.0434 | 1227.9 | 0.0031 | 0.0434 | 0.0347 | | |
| 8 | 90 | 0.0017 | 7832.5 | 0.0039 | 0.0017 | 0.0015 | | |
| 7 | 90 | 0.0240 | 535.3 | 0.0027 | 0.0240 | 0.0192 | | |
| 10 | 120 | 0.0099 | 2121.6 | 0.0033 | 0.0099 | 0.0079 | 0.0164 | 0.0132 |
| 1 | 120 | 0.0087 | 3253.4 | 0.0035 | 0.0087 | 0.007 | | |
| 12 | 120 | 0.0452 | 414.2 | 0.0026 | 0.0452 | 0.0361 | | |
| 6 | 120 | 0.0020 | 10272 | 0.0040 | 0.0020 | 0.0018 | | |
| 0 | 150 | 0.0023 | 3284.6 | 0.0035 | 0.0023 | 0.0020 | 0.0074 | 0.0060 |
| 4 | 150 | 0.0024 | 8034.1 | 0.0039 | 0.0024 | 0.0021 | | |
| 15 | 150 | 0.0147 | 1359.8 | 0.0031 | 0.0147 | 0.0118 | | |
| 11 | 150 | 0.0102 | 2469.9 | 0.0034 | 0.0102 | 0.0082 | | |

Table 5. Average group performance analysis according to the levels of parameter 3 for Sampling Rate $h_3^*$

| Exp. Index | sample Rate $h_3$ | Training error | Train. time t | Log(t)/1000 | Training Obj 1 | Training Obj 2 | average training obj1 | average training obj 2 |
|---|---|---|---|---|---|---|---|---|
| 3 | 0.382 | 0.0967 | 1262.4 | 0.0031 | 0.0967 | 0.0773 | 0.0489 | 0.0391 |
| 14 | 0.382 | 0.0434 | 1227.9 | 0.0031 | 0.0434 | 0.0347 | | |
| 12 | 0.382 | 0.0452 | 414.2 | 0.0026 | 0.0452 | 0.0361 | | |
| 11 | 0.382 | 0.0102 | 2469.9 | 0.0034 | 0.0102 | 0.0082 | | |
| 13 | 0.618 | 0.0373 | 1812.5 | 0.0033 | 0.0373 | 0.0299 | 0.0305 | 0.0245 |
| 5 | 0.618 | 0.0679 | 423.8 | 0.0026 | 0.0679 | 0.0544 | | |
| 15 | 0.618 | 0.0147 | 1359.8 | 0.0031 | 0.0147 | 0.0118 | | |
| 6 | 0.618 | 0.0020 | 10272 | 0.0040 | 0.0020 | 0.0018 | | |
| 10 | 0.8 | 0.0099 | 2121.6 | 0.0033 | 0.0099 | 0.0079 | 0.0119 | 0.0096 |
| 4 | 0.8 | 0.0024 | 8034.1 | 0.0039 | 0.0024 | 0.0021 | | |
| 9 | 0.8 | 0.0113 | 3365.1 | 0.0035 | 0.0113 | 0.0091 | | |
| 7 | 0.8 | 0.0240 | 535.3 | 0.0027 | 0.0240 | 0.0192 | | |
| 0 | 1 | 0.0023 | 3284.6 | 0.0035 | 0.0023 | 0.0020 | 0.0068 | 0.0055 |
| 1 | 1 | 0.0087 | 3253.4 | 0.0035 | 0.0087 | 0.007 | | |
| 8 | 1 | 0.0017 | 7832.5 | 0.0039 | 0.0017 | 0.0015 | | |
| 2 | 1 | 0.0144 | 993.7 | 0.0030 | 0.0144 | 0.0115 | | |

**Table 6.** Average group performance analysis according to the levels of parameter 4 for ResNet $h_4^*$

| Exp. Index | Str. ResNet $h_4$ | Training error | Train. time t | Log(t)/1000 | Training Obj 1 | Training Obj 2 | average training obj1 | average training obj 2 |
|---|---|---|---|---|---|---|---|---|
| 3 | 20 | 0.0967 | 1262.4 | 0.0031 | 0.0967 | 0.0773 | 0.0360 | 0.0288 |
| 1 | 20 | 0.0087 | 3253.4 | 0.0035 | 0.0087 | 0.007 | | |
| 15 | 20 | 0.0147 | 1359.8 | 0.0031 | 0.0147 | 0.0118 | | |
| 7 | 20 | 0.0240 | 535.3 | 0.0027 | 0.0240 | 0.0192 | | |
| 13 | 32 | 0.0373 | 1812.5 | 0.0033 | 0.0373 | 0.0299 | 0.0248 | 0.0199 |
| 4 | 32 | 0.0024 | 8034.1 | 0.0039 | 0.0024 | 0.0021 | | |
| 12 | 32 | 0.0452 | 414.2 | 0.0026 | 0.0452 | 0.0361 | | |
| 2 | 32 | 0.0144 | 993.7 | 0.0030 | 0.0144 | 0.0115 | | |
| 10 | 56 | 0.0099 | 2121.6 | 0.0033 | 0.0099 | 0.0079 | 0.0224 | 0.0180 |
| 5 | 56 | 0.0679 | 423.8 | 0.0026 | 0.0679 | 0.0544 | | |
| 8 | 56 | 0.0017 | 7832.5 | 0.0039 | 0.0017 | 0.0015 | | |
| 11 | 56 | 0.0102 | 2469.9 | 0.0034 | 0.0102 | 0.0082 | | |
| 0 | 110 | 0.0023 | 3284.6 | 0.0035 | 0.0023 | 0.0020 | 0.0148 | 0.0119 |
| 14 | 110 | 0.0434 | 1227.9 | 0.0031 | 0.0434 | 0.0347 | | |
| 9 | 110 | 0.0113 | 3365.1 | 0.0035 | 0.0113 | 0.0091 | | |
| 6 | 110 | 0.0020 | 10272 | 0.0040 | 0.0020 | 0.0018 | | |

**Table 7.** Average group performance analysis according to the levels of parameter 5 for Batch Size $h_5^*$

| Exp. Index | Batch Size $h_5$ | Training error | Train. time t | Log(t)/1000 | Training Obj 1 | Training Obj 2 | average training obj1 | average training obj 2 |
|---|---|---|---|---|---|---|---|---|
| 3 | 32 | 0.0967 | 1262.4 | 0.0031 | 0.0967 | 0.0773 | 0.0257 | 0.0207 |
| 4 | 32 | 0.0024 | 8034.1 | 0.0039 | 0.0024 | 0.0021 | | |
| 8 | 32 | 0.0017 | 7832.5 | 0.0039 | 0.0017 | 0.0015 | | |
| 6 | 32 | 0.0020 | 10272 | 0.0040 | 0.0020 | 0.0018 | | |
| 13 | 64 | 0.0373 | 1812.5 | 0.0033 | 0.0373 | 0.0299 | 0.0169 | 0.0135 |
| 1 | 64 | 0.0087 | 3253.4 | 0.0035 | 0.0087 | 0.007 | | |
| 9 | 64 | 0.0113 | 3365.1 | 0.0035 | 0.0113 | 0.0091 | | |
| 11 | 64 | 0.0102 | 2469.9 | 0.0034 | 0.0102 | 0.0082 | | |
| 10 | 128 | 0.0099 | 2121.6 | 0.0033 | 0.0099 | 0.0079 | 0.0206 | 0.0165 |
| 14 | 128 | 0.0434 | 1227.9 | 0.0031 | 0.0434 | 0.0347 | | |
| 15 | 128 | 0.0147 | 1359.8 | 0.0031 | 0.0147 | 0.0118 | | |
| 2 | 128 | 0.0144 | 993.7 | 0.0030 | 0.0144 | 0.0115 | | |
| 0 | 256 | 0.0023 | 3284.6 | 0.0035 | 0.0023 | 0.0020 | 0.0348 | 0.0279 |
| 5 | 256 | 0.0679 | 423.8 | 0.0026 | 0.0679 | 0.0544 | | |
| 12 | 256 | 0.0452 | 414.2 | 0.0026 | 0.0452 | 0.0361 | | |
| 7 | 256 | 0.0240 | 535.3 | 0.0027 | 0.0240 | 0.0192 | | |

Furthermore, by checking the variation ranges of the average group performance indices for each hyperparameter and each objective case, we may predict the hyperparameter importance rank to the system performance functional variation for each objective case as listed in Table 8 below.

**Table 8. Hyperparameter importance rank to performance variation for objectives**

| Hyperparameters | Average Performance Index Variation Range | | Hyperparameter Rank based on Performance Variation Range | |
|---|---|---|---|---|
| | Obj 1 | Obj 2 | Obj 1 | Obj 2 |
| Learning Rate | 0.0239 | 0.0191 | 3 | 3 |
| Epochs | 0.0402 | 0.0321 | 2 | 2 |
| Sampling Rate | 0.0421 | 0.0336 | 1 | 1 |
| Backbone ResNet | 0.0212 | 0.0169 | 4 | 4 |
| Batch Size | 0.0179 | 0.0144 | 5 | 5 |

For example, the Obj1 variation range of parameter $h_1$ is calculated by the difference between the maximum value and the minimum value in column "average training obj1" as $0.0366 - 0.0127 = 0.0239$. We do similar calculations for parameters $h_2, \cdots, h_5$ based on Tables 4 – 7 respectively as above example from Table 3 for $h_1$.

From Table 8, it is observed that hyperparameter sampling rate ($h_3$) is most important among these five general parameters, and followed by epochs ($h_2$), learning rate ($h_1$), backbone ResNet layers ($h_4$), and batch size ($h_5$). For these two objective cases, their respective hyperparameter importance rank sequences are the same as in Table 8.

From Table 2, it is further noticed that the proposed optimal systems with the searched optimal hyperparameters by the proposed methods have their respective best performances for their respective objectives among these listed orthogonal experiments, not only on the training set but also on the test set, that reflects the system optimality and robustness by the proposed method. The further experiments and research on that are recommended.

We further compare our optimal CNN system model results with the state-of-the-art results in [14][21] with the original backbone ResNet on the CIFAR-10 dataset as listed in Table 9 below.

**Table 9. Comparison of the proposed method results with the reference results in [14][21] on CIFAR-10**

| Method | Training Accuracy | Learn. Rate | Epochs | Sample Rate | Structure backbone | Batch Size | Test Accuracy | Test Error Rate | Total Error Rate (Tra. 50k+ Test 10k) |
|---|---|---|---|---|---|---|---|---|---|
| This paper | 100% | 0.1 | 150 | 1 | Original ResNet 110 | 64 | 94.17% | 5.83% | 0.592% |
| [14] | | | | | Ori. ResNet 110 | | 93.57% | 6.43% | |
| [21] | | | | | Ghost- ResNet 56 | | 92.7% | 7.3% | |

Up to our best knowledge, our resultant optimal CNN system with the backbone original ResNet-110 achieves a best accuracy by the original backbone ResNet [14] on the CIFAR-10 test dataset. From Table 9, it shows that our proposed methods are valued for the fast optimal hyperparameter setting in the AI CNN systems as a demonstration on CIFAR-10 datasets. The proposed optimal system achieves a combination accuracy error rate 0.592% on the CIFAR-10 dataset of 60k samples in view of the error rate 0.010% on the training set of 50k samples and the error rate 5.83% on the test set of 10k samples from Table 2.

## 6 Conclusions

In this paper we propose a new approach and methods to fast set the model hyperparameters for optimal AI systems, especially for CNN systems, via the extended Taguchi approach where their hyperparameters may include the structure selection and their systems can have a single objective or multi-objective. For multi-objective, we take their multi-objective performance functional vector norm as the performance evaluation index. Based on that, we extend the Taguchi method not only for the multiple objective systems, but also for the AI CNN systems and other systems, with the training dataset and the test dataset for training and test respectively as practice.

The proposed approach and methods are based on the effective and efficient Taguchi orthogonal experiments at the selected grid points, which dramatically reduces the number of experiments. It is effective and efficient method for the AI systems and other systems, which have multi-objective optimization tasks.

The proposed methods have been applied to a CNN classification system with the backbone ResNet [14] on the CIFAR-10 database, that also achieves an optimal accuracy for two objective cases. These experiments validate the methods and show the optimal system robustness. We will further test the proposed methods on different datasets.

## Acknowledgement

The first author Sheng-Guo Wang expresses his special thanks to LeiLani Paugh and Morgan Weatherford at NCDOT for their support to his research.